\documentclass[preprint,authoryear,12pt]{elsarticle}

\usepackage{amssymb}
\usepackage{graphicx}
\usepackage{booktabs}
\usepackage{tabularx}
\usepackage{array}
\usepackage{lscape}
\usepackage{longtable}
\usepackage{epstopdf}
\usepackage{graphicx}    
\usepackage{booktabs}    
\usepackage{colortbl}    
\usepackage{threeparttable}
\usepackage{rotating}

\newcommand{\PreserveBackslash}[1]{\let\temp=\\#1\let\\=\temp}
\newcolumntype{C}[1]{>{\PreserveBackslash\centering}p{#1}}
\newcolumntype{R}[1]{>{\PreserveBackslash\raggedleft}p{#1}}
\newcolumntype{L}[1]{>{\PreserveBackslash\raggedright}p{#1}}
\newtheorem{definition}{Definition}[section]

\journal{arXiv.org}

\begin{document}

\begin{frontmatter}



\title{Ranking basic belief assignments in decision making under uncertain environment}


\author[address1]{Yuxian Du}
\author[address1]{Shiyu Chen}
\author[address2]{Yong Hu}
\author[address3]{Felix T.S. Chan}
\author[address4]{Sankaran Mahadevan}
\author[address1,address4]{Yong Deng\corref{label1}}

\cortext[label1]{Corresponding author: Yong Deng, School of Computer and Information Science, Southwest University, Chongqing, 400715, China. Email address: ydeng@swu.edu.cn; prof.deng@hotmail.com. Tel.: +86 23 6825 4555, Fax: +86 23 6825 4555}

\address[address1]{School of Computer and Information Science, Southwest University, Chongqing 400715, China}
\address[address2]{Institute of Business Intelligence and Knowledge Discovery, Guangdong University of Foreign Studies, Guangzhou 510006, China}
\address[address3]{Department of Industrial and Systems Engineering, the Hong Kong Polytechnic University, Hong Kong, China}
\address[address4]{School of Engineering, Vanderbilt University, Nashville, TN 37235, USA}

\begin{abstract}
Dempster-Shafer theory (D-S theory) is widely used in decision making under the uncertain environment. Ranking basic belief assignments (BBAs) now is an open issue. Existing evidence distance measures cannot rank the BBAs in the situations when the propositions have their own ranking order or their inherent measure of closeness. To address this issue, a new ranking evidence distance (RED) measure is proposed. Compared with the existing evidence distance measures including the Jousselme's distance and  the distance between betting commitments, the proposed RED measure is much more general due to the fact that the order of the propositions in the systems is taken into consideration. If there is no order or no inherent measure of closeness in the propositions, our proposed RED measure is reduced to the existing evidence distance. Numerical examples show that the proposed RED measure is an efficient alternative to rank BBAs in decision making under uncertain environment.
\end{abstract}


\end{frontmatter}


\section{Introduction}\label{Introduction}
With the ability of dealing with the uncertainty or imprecision embedded in evidence, the Dempster-Shafer theory (D-S theory), which was first proposed by \cite{dempster1967upper} and then developed by \cite{shafer1976mathematical}, is widely used in many applications. Ranking basic belief assignments (BBAs) in D-S theory now is an open issue. How can we know whether a BBA is ``far'' from the solution or ``close'' to it? Once this ``distance'' to the solution is quantified, one is able to observe the progress the algorithm makes as it converges on a solution.

\subsection{Previous work}
Several works could be found in the literatures which tried to propose distance measures through the evidence theory frame work \citep{denoeux2001handling,khatibi2010new,jousselme2001new,liu2006analyzing}. \cite{denoeux2001handling} proposed a Euclidean measure for the evidence theory with taking into account partial knowledge of the class of training samples. \cite{khatibi2010new} proposed a new evidential distance measure based on belief intervals, which these functions were according to nearest neighborhood concept. \cite{jousselme2001new} applied a classical similarity measure to achieve the comparison of the focal elements of two BBAs, in order to define a distance in a vector space generated by the focal elements. \cite{liu2006analyzing} investigated the meaning of pignistic transformation and defined the distance between betting commitments from two pignistic transformations.

\subsection{Problem description}
Though a large number of researches about evidence distance are proposed, there are still some situations which cannot be solved. The existing evidence distance measures \citep{jousselme2001new,liu2006analyzing} cannot rank the BBAs in the situations when the propositions have their own ranking or their inherent measure of closeness. This issue is described as an example as below.

\textbf{\textit{Example 1 }}In this paper, we choose linguistic variables for the assessment of risk factors, and the individual evaluation grade is defined as $\{ Poor, Low, Middle, High, Perfect\} $. The frame of discernment $\{ 1,2,3,4,5\} $ is expressed as the linguistic variables $\{ Poor, Low, Middle, High, Perfect\} $, respectively. Let the BBAs are:
\[{m_1}:{m_1}(\{ 1\} ) = 1;~~~{\rm{    }}
{m_2}:{m_2}(\{ 2\} ) = 1;~~~{\rm{    }}
{m_3}:{m_3}(\{ 5\} ) = 1.\]
The Jousselme's evidence distance \citep{jousselme2001new} is calculated:
\[d_{BBA}^J({m_1},{m_2}) = 1,~~d_{BBA}^J({m_1},{m_3}) = 1\]
The distance between betting commitments \citep{liu2006analyzing} is calculated:
\[d_{BBA}^{PPT}({m_1},{m_2}) = 1,~d_{BBA}^{PPT}({m_1},{m_3}) = 1\]

Both in the two existing evidence distance measures, the distance between $m_1$ and $m_2$ is as same as that between $m_1$ and $m_3$. In this case, we cannot rank the BBAs in order with this distance. However, in many real applications, the distance of the assessment between ``$Poor$'' and ・・$Low$'' would be smaller than that between ``$Poor$'' and ``$Middle$'', which means that the assessment with ``$Low$'' is closer to ``$Poor$'' than ``$Middle$''. Aiming at this issue, a new ranking evidence distance (RED) measure with correlation matrix is proposed.

The rest of the paper is organized as follows. Section 2 introduces some preliminaries about Dempster-Shafer theory, pignistic probability transformation and evidence distance measures. The new ranking evidence distance (RED) measure with correlation matrix is proposed in section 3. Numerical examples are given to show the efficiency of the proposed evidence distance in section 4. A short conclusion is drawn in the section 5.

\section{Preliminaries}

\subsection{Dempster-Shafer theory}
Dempster-Shafer evidence theory as an important method has been widely used in many fields such as decision making, failure detection, information fusion and so on \citep{beynon2005novel,yang2006evidential,yang2009integrating,Deng2011Modeling,beynon2011introduction
}.

\begin{definition}
Let $\Theta$ be a finite nonempty set of mutually exclusive and exhaustive hypotheses, called the frame of discernment, where $\Theta  = \{ 1,2, \ldots ,N\}$. A mass function is a mapping $m:2^{\Theta} \to [0,1]$, which satisfies:
\begin{equation}\label{m0}
m(\emptyset ) = 0~~~{\rm{    }}and~~~{\rm{    }}
\sum\limits_{A \subseteq 2^{\Theta}} {m(A)}  = 1
\end{equation}
\end{definition}

Note that $2^\Theta$ is an exhaustive set that contains $2^N$ elements. If $m(A) > 0$, $A$ is called a focal element.

Two BBAs $m_1$ and $m_2$ (i.e., two bodies of evidence or two belief functions) can be combined to yield a new BBA $m$, by Dempster's combination rule \cite{dempster1967upper}.

\begin{definition}
Dempster's rule of combination, denoted by $({m_1} \oplus {m_2})$, called orthogonal sum of $m_1$ an $m_2$, is defined as follows:
\begin{equation}
m(A) = {m_1}(A) \oplus {m_2}(A) = \frac{{\sum\nolimits_{B \cap C = A} {{m_1}(B){m_2}(C)} }}{{1 - k}}
\end{equation}
with
\begin{equation}
k = \sum\limits_{B \cap C = \emptyset } {{m_1}(B){m_2}(C)}
\end{equation}
where $K$ is a normalization constant, called conflict because it measures the degree of conflict between $m_1$ and $m_2$.
\end{definition}
\subsection{Pignistic probability transformation}
\begin{definition}\label{defPPT}
Let m be a BBA on $\Theta$. The resulting pignistic probability transformation (PPT) for the singletons $x \in \Theta $ is given by \cite{smets1994transferable}:
\begin{equation}\label{PPT}
BetP(\{ x\} ) = \sum\limits_{x \in A \subseteq \Theta } {\frac{1}{{\left| A \right|}}\frac{{m(A)}}{{1 - m(\emptyset )}}} ,~~~{\rm{    }}m(\emptyset ) \ne 1
\end{equation}
\end{definition}
where ${\left| A \right|}$ is the number of elements of $\Theta$ in A. For non-singleton $x \subseteq \Theta $,we have:
\begin{equation}
BetP(A) = \sum\limits_{x \in A} {BetP(\{ x\} )}
\end{equation}

\begin{definition}
Let $m_1$ and $m_2$ be two BBAs defined on the same frame of discernment $\Theta$. And Let $BetP_{m_1}$ and $BetP_{m_2}$ be the results of two pignistic transformations from them respectively. Then
\begin{equation}
difBetP_{{m_1}}^{{m_2}} = {\max _{A \subseteq \Theta }}(|Bet{P_{{m_1}}}(A) - Bet{P_{{m_2}}}(A)|)
\end{equation}
is called the distance between betting commitments of the two BBAs \citep{liu2006analyzing}. For simplicity, here we set $d_{BBA}^{PPT}({m_1},{m_2}) = difBetP_{{m_1}}^{{m_2}}$.
\end{definition}

\subsection{Evidential distance measure}

\begin{definition}\label{evidence distance}
Let $m_1$ and $m_2$ be two BBAs on the same frame of discernment $\Theta$, containing $N$ mutually exclusive and exhaustive hypotheses. The distance between $m_1$ and $m_2$ by \cite{jousselme2001new} is:
\begin{equation}\label{d_{BBA}}
{d_{BBA}^J}({m_1},{m_2}) = \sqrt {\frac{1}{2}{{({{\overrightarrow m }_1} - {{\overrightarrow m }_2})}^{\rm{T}}}\underline{\underline D} ({{\overrightarrow m }_1} - {{\overrightarrow m }_2})}
\end{equation}
where ${{{\overrightarrow m }_1}}$ and ${{{\overrightarrow m }_2}}$ are ``$mass~vectors$'' whoes elements are the masses corresponding to each of the members of the combined set of focal elements from both BBAs. ${\underline{\underline D} }$ is an ${2^N} \times {2^N}$ matrix whose elements are $D(A,B) = \frac{{\left| {A \cap B} \right|}}{{\left| {A \cup B} \right|}}$, $A,B \in P(\Theta )$.
\end{definition}

\section{Proposed method}

As mentioned above, existing evidence distance measures cannot rank the BBAs in the situations when the propositions have their own ranking order or their inherent measure of closeness. To address this issue as mentioned in the above example, a new ranking evidence distance (RED) measure with correlation matrix is proposed. If there is no order or no inherent measure of closeness in the propositions, our proposed RED measure is reduced to the Jousselme's evidence distance \citep{jousselme2001new}.

\subsection{RED definition}
\begin{definition}\label{RED}
Let $m_1$ and $m_2$ be two BBAs on the same frame of discernment $\Theta$, containing $N$ mutually exclusive and exhaustive hypotheses. The distance between $m_1$ and $m_2$ is:
\begin{equation}\label{d_{BBA}}
{d_{BBA}^{RED}}({m_1},{m_2}) = \sqrt {\frac{1}{2}{{({{\vec m}_1} - {{\vec m}_2})}^{\rm{T}}}\underline{\underline D}\; \underline{\underline S} ({{\vec m}_1} - {{\vec m}_2})}
\end{equation}
where ${{{\overrightarrow m }_1}}$ and ${{{\overrightarrow m }_2}}$ are the BBAs, ${\underline{\underline D} }$ and ${\underline{\underline S} }$ are two $N \times N$ matrices, and elements in ${\underline{\underline D} }$ are $D(A,B) = \frac{{\left| {A \cap B} \right|}}{{\left| {A \cup B} \right|}}$, $A,B \in P(\Theta )$.
\end{definition}

PPT 优势、意义、理由

In Definition \ref{RED}, due to that fact we apply PPT to effectively transform the BBA with non-singlet subsets focal elements to the new BBA with only singlet focal elements, the number of singlet focal elements is equal to the number of the members of the frame of discernment. So, the dimension of ${\underline{\underline D} }$ is $N \times N$. Further more, ${\underline{\underline D} }$ is evolved into the unit matrix:
\[\underline{\underline D} = \left[ {\begin{array}{*{20}{c}}
1&0& \cdots &0\\
0&1& \cdots &0\\
 \vdots & \vdots & \ddots & \vdots \\
0&0& \cdots &1
\end{array}} \right]\]

To illustrate ${\underline{\underline D} }$ in the Definition \ref{RED}, a example is given, defined $m_1$, over a frame of discernment $\Theta  = \{ {x_1},{x_2},{x_3}\}$:
\[{m_1}:~~~{\rm{    }}{m_1}(\{ {x_1}\} ) = 0.3,~{m_1}(\{ {x_1},{x_2}\} ) = 0.4,~{m_1}(\{ {x_1},{x_2},{x_3}\} ) = 0.3\]
The new BBAs is calculated by using PPT in Definition \ref{defPPT}:
\[m_1^{'}(\{ {x_1}\} ) = {m_1}(\{ {x_1}\} ) + \frac{{{m_1}(\{ {x_1},{x_2}\} )}}{2} + \frac{{{m_1}(\{ {x_1},{x_2},{x_3}\} )}}{3} = 0.3 + \frac{{0.4}}{2} + \frac{{0.3}}{3} = 0.6\]
\[m_1^{'}(\{ {x_2}\} ) = \frac{{{m_1}(\{ {x_1},{x_2}\} )}}{2} + \frac{{{m_1}(\{ {x_1},{x_2},{x_3}\} )}}{3} = \frac{{0.4}}{2} + \frac{{0.3}}{3} = 0.3\]
\[m_1^{'}(\{ {x_3}\} ) = \frac{{{m_1}(\{ {x_1},{x_2},{x_3}\} )}}{3} = \frac{{0.3}}{3} = 0.1\]
\[{m_1^{'}}:~~~{\rm{    }}{m_1}(\{ {x_1}\} ) = 0.6,~{m_1}(\{ {x_2}\} ) = 0.2,~{m_1}(\{ {x_3}\} ) = 0.1\]
Due to $D(A,B) = \frac{{\left| {A \cap B} \right|}}{{\left| {A \cup B} \right|}}$, $A,B \in P(\Theta )$, ${\underline{\underline D} }$ is calculated:
\[\begin{array}{*{20}{c}}
{{d_{ij}}}&{\left\{ {{x_1}} \right\}}&{\left\{ {{x_2}} \right\}}&{\left\{ {{x_3}} \right\}}\\
{\left\{ {{x_1}} \right\}}&1&0&0\\
{\left\{ {{x_2}} \right\}}&0&1&0\\
{\left\{ {{x_3}} \right\}}&0&0&1
\end{array}\begin{array}{*{20}{c}}
~~~{\rm{    }}~~~{\rm{    }}~~~{\rm{    }}~~~{\rm{    }}{ \underline{\underline D} = \left[ {\begin{array}{*{20}{c}}
1&0&0\\
0&1&0\\
0&0&1
\end{array}} \right]}
\end{array}\]
So, if the number of the members of the frame of discernment is $N$, the dimension of ${\underline{\underline D} }$ of the new BBA is $N \times N$.

\begin{definition}
$\underline{\underline S}  = ({s_{ij}})$, is called correlation matrix. The value of ${s_{ij}}$ represents the degree of correlation (or closeness) between focal elements $i$ and $j$. It is defined as :
\begin{equation}\label{s_{ij}}
{s_{ij}} = 1 - \left| {i - j} \right|\frac{1}{{N - 1}}
\end{equation}
\end{definition}
with $i = 1, \cdots ,N$ and $j = 1, \cdots ,N$, where $N$ is the number of elements of the frame of discernment. $\underline{\underline S}$ can be obtained by
\begin{equation}
\underline{\underline S} = \left[ {\begin{array}{*{20}{c}}
1&{1 - \frac{2}{{N - 1}}}&{}& \cdots &{}&{1 - \frac{{N - 2}}{{N - 1}}}&{1 - \frac{{N - 1}}{{N - 1}}}\\
{1 - \frac{1}{{N - 1}}}&1&{}& \cdots &{}&{1 - \frac{{N - 3}}{{N - 1}}}&{1 - \frac{{N - 2}}{{N - 1}}}\\
{}&{}&{}&{}&{}&{}&{}\\
 \vdots &{}&{}& \ddots &{}&{}& \vdots \\
{}&{}&{}&{}&{}&{}&{}\\
{1 - \frac{{N - 2}}{{N - 1}}}&{1 - \frac{{N - 3}}{{N - 1}}}&{}& \cdots &{}&1&{1 - \frac{1}{{N - 1}}}\\
{1 - \frac{{N - 1}}{{N - 1}}}&{1 - \frac{{N - 2}}{{N - 1}}}&{}& \cdots &{}&{1 - \frac{1}{{N - 1}}}&1
\end{array}} \right]
\end{equation}

For example, in risk evaluation, we employe linguistic variables for the assessment, and the individual evaluation grade is defined as
\[H = \{ 1,2,3,4,5\}  = \{ Poor,Low,Middle,High,Perfect\} \]

Due to $i = 1, \cdots ,5$ and $j = 1, \cdots ,5$, $s_{11}$=$s_{22}$=$s_{33}$=$s_{44}$=$s_{55}$=1. That is to say the degree of correlation is perfect correlation and identical for itself. $\frac{1}{{N - 1}}$ represents the difference degree between the neighbouring evaluation grade, such as between ``$Poor$'' and ``$Low$'', between ``$Low$'' and ``$Middle$'', between ``$Middle$'' and ``$High$'', and between ``$High$'' and ``$Perfect$''. So the correlation matrix $\underline{\underline S} $ is obtained by:
\begin{table}[!ht]
\centering
\caption{Correlation matrix}\label{table sij}
\begin{tabular}{c|ccccc}
  ${s_{ij}}$ & 1 & 2 & 3 & 4 & 5 \\
  \hline
  1 & 1 & 0.75 & 0.5 & 0.25 & 0 \\
  2 & 0.75 & 1 & 0.75 & 0.5 & 0.25 \\
  3 & 0.5 & 0.75 & 1 & 0.75 & 0.5 \\
  4 & 0.25 & 0.5 & 0.75 & 1 & 0.75 \\
  5 & 0 & 0.25 & 0.5 & 0.75 & 1 \\
\end{tabular}
\end{table}

%
\begin{equation}\label{S}
\underline{\underline S}  = \left[ {\begin{array}{*{20}{c}}
1&{0.75}&{0.5}&{0.25}&0\\
{0.75}&1&{0.75}&{0.5}&{0.25}\\
{0.5}&{0.75}&1&{0.75}&{0.5}\\
{0.25}&{0.5}&{0.75}&1&{0.75}\\
0&{0.25}&{0.5}&{0.75}&1
\end{array}} \right]
\end{equation}


Without loss of generality, if there is no order or no inherent measure of closeness in the propositions, our proposed RED is reduced to the Jousselme's evidence distance \citep{jousselme2001new}. In this case, the correlation matrix $\underline{\underline S}$ is a unit matrix:
\[\underline{\underline S} = \left[ {\begin{array}{*{20}{c}}
1&0& \cdots &0\\
0&1& \cdots &0\\
 \vdots & \vdots & \ddots & \vdots \\
0&0& \cdots &1
\end{array}} \right]\]
It means that there is no difference among these evaluation grades in the aspect of order or inherent measure of closeness. In this situation, the proposed new RED with correlation matrix is reduced to Jousselme's evidence distance \citep{jousselme2001new}:
\begin{equation}
\begin{array}{l}
d_{BBA}^{RED}({m_1},{m_2}) = \sqrt {\frac{1}{2}{{({{\vec m}_1} - {{\vec m}_2})}^{\rm{T}}}\underline{\underline D}\; \underline{\underline S}({{\vec m}_1} - {{\vec m}_2})} \\
~~~{\rm{    }}~~~{\rm{    }}~~~{\rm{    }}~~~{\rm{    }}~~~{\rm{    }}~~~{\rm{    }} = \sqrt {\frac{1}{2}{{({{\vec m}_1} - {{\vec m}_2})}^{\rm{T}}}\underline{\underline D}({{\vec m}_1} - {{\vec m}_2})} \\
~~~{\rm{    }}~~~{\rm{    }}~~~{\rm{    }}~~~{\rm{    }}~~~{\rm{    }}~~~{\rm{    }} ={d_{BBA}^J}({m_1},{m_2})
\end{array}
\end{equation}

\subsection{RED properties}
The proposed RED

\section{Numerical examples}

To illustrated the effectiveness of the proposed method, three examples with different BBAs are performed in this section, which compares the proposed method with the existing evidence distance measures \citep{jousselme2001new,liu2006analyzing}.

In all three examples, suppose the frame of discernment $\{ 1,2,3,4,5\} $ is expressed as $\{ Poor, Low, Middle, High, Perfect\} $, respectively.

\textbf{\textit{Example 2 }}Let the BBAs be:
\[\begin{array}{l}
{m_1}:~~~{\rm{    }}{m_1}(\{ 1\} ) = 1;\\
{m_2}:~~~{\rm{    }}{m_2}(\{ 2\} ) = 1;\\
{m_3}:~~~{\rm{    }}{m_3}(\{ 3\} ) = 1.
\end{array}\]

$({{\vec m}_1} - {{\vec m}_2})$, $({{\vec m}_1} - {{\vec m}_3})$ and ${\underline{\underline D} }$ are obtained:
\[({\vec m_1} - {\vec m_2}) = \left( {\begin{array}{*{20}{c}}
1\\
{ -1}
\end{array}} \right),~~~{\rm{    }}({\vec m_1} - {\vec m_3}) = \left( {\begin{array}{*{20}{c}}
1\\
{ -1}
\end{array}} \right)\]
\[\underline{\underline D} ({m_1},{m_2}) = \left[ {\begin{array}{*{20}{c}}
1&0\\
0&1
\end{array}} \right],~~~{\rm{    }}\underline{\underline D} ({m_1},{m_3}) = \left[ {\begin{array}{*{20}{c}}
1&0\\
0&1
\end{array}} \right]\]

According to Eq.(\ref{s_{ij}}), the correlation matrix $\underline{\underline S} $ is obtained from Table \ref{table sij}:
\[\underline{\underline S} {({m_1},{m_2}) = \left[ {\begin{array}{*{20}{c}}
1&{0.75}\\
{0.75}&1
\end{array}} \right]},~~~{\rm{    }}\underline{\underline S} {({m_1},{m_3}) = \left[ {\begin{array}{*{20}{c}}
1&0.5\\
0.5&1
\end{array}} \right]}\]

So, the new RED ${d_{BBA}^{RED}}$ is calculated by Eq.(\ref{d_{BBA}}):
\[d_{BBA}^{RED}({m_1},{m_2}) = \sqrt {\frac{1}{2}\left( {1, - 1} \right)\left( {\begin{array}{*{20}{c}}
1&0\\
0&1
\end{array}} \right)\left( {\begin{array}{*{20}{c}}
1&{0.75}\\
{0.75}&1
\end{array}} \right)\left( {\begin{array}{*{20}{c}}
1\\
{ - 1}
\end{array}} \right)}  = 0.5\]

\[d_{BBA}^{RED}({m_1},{m_3}) = \sqrt {\frac{1}{2}\left( {1, - 1} \right)\left( {\begin{array}{*{20}{c}}
1&0\\
0&1
\end{array}} \right)\left( {\begin{array}{*{20}{c}}
1&0.5\\
0.5&1
\end{array}} \right)\left( {\begin{array}{*{20}{c}}
1\\
{ - 1}
\end{array}} \right)}  = 0.707\]


%

\textbf{\textit{Example 3 }}Let the BBAs be:
\[{m_1}:{m_1}(\{ 1\} ) = 1;~~~{\rm{    }}
{m_2}:{m_2}(\{ 2,3\} ) = 1;~~~{\rm{    }}
{m_3}:{m_3}(\{ 4,5\} ) = 1.\]

We apply pignistic probability transformation (PPT) by \cite{smets1994transferable} to transform the BBAs with non-singlet focal elements to the BBAs with only singlet focal elements by Eq.(\ref{PPT}):
\[\begin{array}{l}
{m_1}:~~~{\rm{    }}{m_1}(\{ 1\} ) = 1;\\
{m_2}:~~~{\rm{    }}{m_2}(\{ 2\} ) = 0.5,~{m_2}(\{ 3\} ) = 0.5;\\
{m_3}:~~~{\rm{    }}{m_3}(\{ 4\} ) = 0.5,~{m_3}(\{ 5\} ) = 0.5.
\end{array}\]

$({{\vec m}_1} - {{\vec m}_2})$, $({{\vec m}_1} - {{\vec m}_3})$ and ${\underline{\underline D} }$ are obtained:
\[({\vec m_1} - {\vec m_2}) = \left( {\begin{array}{*{20}{c}}
1\\
{ - 0.5}\\
{ - 0.5}
\end{array}} \right),~~~{\rm{    }}({\vec m_1} - {\vec m_3}) = \left( {\begin{array}{*{20}{c}}
1\\
{ - 0.5}\\
{ - 0.5}
\end{array}} \right)\]
\[\underline{\underline D} ({m_1},{m_2}) = \left[ {\begin{array}{*{20}{c}}
1&0&0\\
0&1&0\\
0&0&1
\end{array}} \right],~~~{\rm{    }}\underline{\underline D} ({m_1},{m_3}) = \left[ {\begin{array}{*{20}{c}}
1&0&0\\
0&1&0\\
0&0&1
\end{array}} \right]\]

According to Eq.(\ref{s_{ij}}), the correlation matrix $\underline{\underline S} $ is obtained from Table \ref{table sij}:
\[\underline{\underline S} ({m_1},{m_2}) = \left[ {\begin{array}{*{20}{c}}
1&0.75&0.5\\
0.75&1&0.75\\
0.5&0.75&1
\end{array}} \right],~~~{\rm{    }}\underline{\underline S} ({m_1},{m_3}) = \left[ {\begin{array}{*{20}{c}}
1&0.25&0\\
0.25&1&0.75\\
0&0.75&1
\end{array}} \right]\]

So, the new RED ${d_{BBA}^{RED}}$ is calculated by Eq.(\ref{d_{BBA}}):
\[d_{BBA}^{RED}({m_1},{m_2}) = \sqrt {\frac{1}{2}\left( {1, - 0.5, - 0.5} \right)\left( {\begin{array}{*{20}{c}}
1&0&0\\
0&1&0\\
0&0&1
\end{array}} \right)\left( {\begin{array}{*{20}{c}}
1&{0.75}&{0.5}\\
{0.75}&1&{0.75}\\
{0.5}&{0.75}&1
\end{array}} \right)\left( {\begin{array}{*{20}{c}}
1\\
{ - 0.5}\\
{ - 0.5}
\end{array}} \right)}  = 0.559\]

\[d_{BBA}^{RED}({m_1},{m_2}) = \sqrt {\frac{1}{2}\left( {1, - 0.5, - 0.5} \right)\left( {\begin{array}{*{20}{c}}
1&0&0\\
0&1&0\\
0&0&1
\end{array}} \right)\left( {\begin{array}{*{20}{c}}
1&{0.25}&{0}\\
{0.25}&1&{0.75}\\
{0}&{0.75}&1
\end{array}} \right)\left( {\begin{array}{*{20}{c}}
1\\
{ - 0.5}\\
{ - 0.5}
\end{array}} \right)}  = 0.901\]


%

\textbf{\textit{Example 4 }}Let the BBAs be:
\[{m_1}:{m_1}(\{ 1\} ) = 1;~~~{\rm{    }}
{m_2}:{m_2}(\{ 1,2\} ) = 1;~~~{\rm{    }}
{m_3}:{m_3}(\{ 1,3\} ) = 1.\]

Pignistic probability transformation (PPT) by \cite{smets1994transferable} is applied by Eq.(\ref{PPT}):
\[\begin{array}{l}
{m_1}:~~~{\rm{    }}{m_1}(\{ 1\} ) = 1;\\
{m_2}:~~~{\rm{    }}{m_2}(\{ 1\} ) = 0.5,~{m_2}(\{ 2\} ) = 0.5;\\
{m_3}:~~~{\rm{    }}{m_3}(\{ 1\} ) = 0.5,~{m_3}(\{ 3\} ) = 0.5.
\end{array}\]

$({{\vec m}_1} - {{\vec m}_2})$, $({{\vec m}_1} - {{\vec m}_3})$ and ${\underline{\underline D} }$ are obtained:
\[({\vec m_1} - {\vec m_2}) = \left( {\begin{array}{*{20}{c}}
0.5\\
{ -0.5}
\end{array}} \right),~~~{\rm{    }}({\vec m_1} - {\vec m_3}) = \left( {\begin{array}{*{20}{c}}
0.5\\
{ -0.5}
\end{array}} \right)\]
\[\underline{\underline D} ({m_1},{m_2}) = \left[ {\begin{array}{*{20}{c}}
1&0.5\\
0.5&1
\end{array}} \right],~~~{\rm{    }}\underline{\underline D} ({m_1},{m_3}) = \left[ {\begin{array}{*{20}{c}}
1&0.5\\
0.5&1
\end{array}} \right]\]

According to Eq.(\ref{s_{ij}}), the correlation matrix $\underline{\underline S} $ is obtained from Table \ref{table sij}:
\[\underline{\underline S} {({m_1},{m_2}) = \left[ {\begin{array}{*{20}{c}}
1&{0.75}\\
{0.75}&1
\end{array}} \right]},~~~{\rm{    }}\underline{\underline S} {({m_1},{m_3}) = \left[ {\begin{array}{*{20}{c}}
1&0.5\\
0.5&1
\end{array}} \right]}\]

So, the new RED ${d_{BBA}^{RED}}$ is calculated by Eq.(\ref{d_{BBA}}):
\[d_{BBA}^{RED}({m_1},{m_2}) = \sqrt {\frac{1}{2}\left( {0.5, -0.5} \right)\left( {\begin{array}{*{20}{c}}
1&0\\
0&1
\end{array}} \right)\left( {\begin{array}{*{20}{c}}
1&{0.75}\\
{0.75}&1
\end{array}} \right)\left( {\begin{array}{*{20}{c}}
0.5\\
{ -0.5}
\end{array}} \right)}  = 0.25\]

\[d_{BBA}^{RED}({m_1},{m_2}) = \sqrt {\frac{1}{2}\left( {0.5, -0.5} \right)\left( {\begin{array}{*{20}{c}}
1&0\\
0&1
\end{array}} \right)\left( {\begin{array}{*{20}{c}}
1&{0.5}\\
{0.5}&1
\end{array}} \right)\left( {\begin{array}{*{20}{c}}
0.5\\
{ -0.5}
\end{array}} \right)}  = 0.354\]

\begin{table}[!h]
{\footnotesize%
\caption{Comparison of the distance results by different distance measures}\label{centralities}
\begin{tabular*}{\columnwidth}{@{\extracolsep{\fill}}@{~~}lllllll@{~~}}
\toprule
   \multicolumn{1}{l}{Method}&\multicolumn{2}{l}{$Example~2$}&\multicolumn{2}{l}{$Example~3$}&\multicolumn{2}{l}{$Example~4$}  \\
   \cmidrule(r){2-3}\cmidrule(r){4-5}\cmidrule(r){6-7}
   \multicolumn{ 1}{l}{}   & $({m_1},{m_2})$ & $({m_1},{m_3})$ & $({m_1},{m_2})$ & $({m_1},{m_3})$ & $({m_1},{m_2})$ & $({m_1},{m_3})$ \\
\midrule
  $d_{BBA}^J$      & 1   & 1     & 1     & 1     & 1    & 1        \\
  $d_{BBA}^{PPT}$  & 1   & 1     & 1     & 1     & 0.5  & 0.5      \\
  $d_{BBA}^{RED}$    & 0.5 & 0.707 & 0.559 & 0.901 & 0.25 & 0.354    \\

\bottomrule
\end{tabular*}
}
\end{table}

The BBAs with smaller distance to the reference substance $m_1$, should be ranked to be more higher. In these examples, we can easily obtain the distances in different methods between $m_1$ and $m_1$ ${d_{BBA}^{R}}({m_1},{m_1}) = {d_{BBA}^J}({m_1},{m_1}) = {d_{BBA}^{PPT}}({m_1},{m_1}) = 0$. And the existing evidence distance measures by \cite{jousselme2001new} and \cite{liu2006analyzing} between $m_1$ and $m_2$ are as the same as between $m_1$ and $m_3$. The existing evidence distance measures \citep{jousselme2001new,liu2006analyzing} could not distinguish the evidence distances and cannot rank the BBAs in all three examples, while the proposed method can obtain $d_{BBA}^R({m_1},{m_1}) < d_{BBA}^R({m_1},{m_2}) < d_{BBA}^R({m_1},{m_3})$ in all three examples. It means that the ranking of the BBAs is ${m_1} \succ {m_2} \succ {m_3}$ in all three examples, where ``$\succ$'' represents ``$better~than$''. With our proposed new RED, we can effectively rank the BBAs, which have a natural order or a inherent measure of closeness.

\begin{table}[!h]
{\footnotesize
\caption{Comparsion of $d_{BBA}^{RED}$, $d_{BBA}^J$ and $d_{BBA}^{PPT}$ of $m_1$ and $m_2$ when subset $A$ changes.}\label{A}
\begin{tabular*}{\columnwidth}{@{\extracolsep{\fill}}@{~~}llll@{~~}}
\toprule
  Cases                     & $d_{BBA}^J$ & $d_{BBA}^{PPT}$ & $d_{BBA}^{RED}$ \\
\midrule
  $A = \{ 1\}$              & 0.7858      & 0.605           & 0.1871      \\
  $A = \{ 1,2\}$            & 0.6866      & 0.426           & 0.1340      \\
  $A = \{ 1,2,3\}$          & 0.5633      & 0.248           & 0.0882      \\
  $A = \{ 1, \ldots ,4\}$   & 0.4286      & 0.125           & 0.0555      \\
  $A = \{ 1, \ldots ,5\}$   & 0.1322      & 0.125           & 0.0597      \\
  $A = \{ 1, \ldots ,6\}$   & 0.3883      & 0.258           & 0.0969      \\
  $A = \{ 1, \ldots ,7\}$   & 0.5029      & 0.355           & 0.1349      \\
  $A = \{ 1, \ldots ,8\}$   & 0.5705      & 0.425           & 0.1682      \\
  $A = \{ 1, \ldots ,9\}$   & 0.6187      & 0.480           & 0.1980      \\
  $A = \{ 1, \ldots ,10\}$  & 0.6553      & 0.525           & 0.2251      \\
  $A = \{ 1, \ldots ,11\}$  & 0.6844      & 0.560           & 0.2499      \\
  $A = \{ 1, \ldots ,12\}$  & 0.7081      & 0.591           & 0.2728      \\
  $A = \{ 1, \ldots ,13\}$  & 0.7274      & 0.617           & 0.2943      \\
  $A = \{ 1, \ldots ,14\}$  & 0.7444      & 0.639           & 0.3144      \\
  $A = \{ 1, \ldots ,15\}$  & 0.7592      & 0.658           & 0.3333      \\
  $A = \{ 1, \ldots ,16\}$  & 0.7658      & 0.675           & 0.3512      \\
  $A = \{ 1, \ldots ,17\}$  & 0.7839      & 0.689           & 0.3682      \\
  $A = \{ 1, \ldots ,18\}$  & 0.7944      & 0.702           & 0.3844      \\
  $A = \{ 1, \ldots ,19\}$  & 0.8042      & 0.714           & 0.3999      \\
  $A = \{ 1, \ldots ,20\}$  & 0.8123      & 0.725           & 0.4147      \\
\bottomrule
\end{tabular*}
}
\end{table}

\begin{figure}
\label{distance_A}
\centering
\includegraphics[scale=0.5]{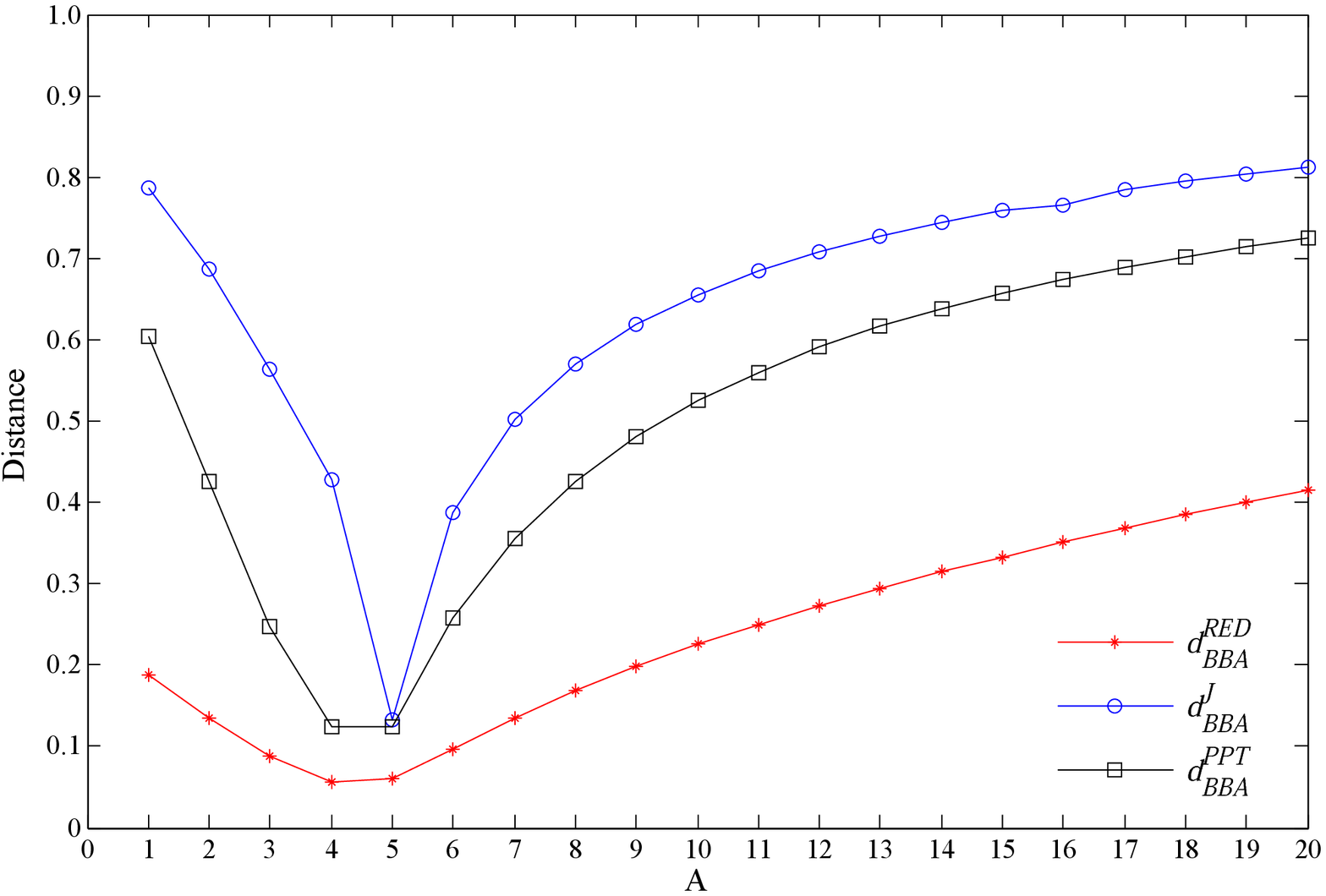}
 \caption{Comparison of different evidence distance measures of the two BBAs detailed in Table when subset $A$ changes. The X-axis shows the sizes of subset $A$ and the Y-axis gives the scale of $d_{BBA}^{RED}$, $d_{BBA}^J$ and $d_{BBA}^{PPT}$.}
\end{figure}

\section{Conclusions}

Ranking alternatives is a very important step in decision making. In some situations based on evidence theory, it is necessary to develop a distance function to rank BBAs. In this paper, a new ranking evidential distance (RED) is proposed. Numerical examples show that the proposed new evidential distance measure is more general to distinguish the distance. With our proposed new RED, the evidence distance can be expressed in these situations to rank BBAs in decision making under uncertain environment. If there is no order or no inherent measure of closeness in the propositions, our proposed RED is reduced to the existing evidence distance.

\section{Acknowledgment}
The work is partially supported by National Natural Science Foundation of China, Grant Nos. 61174022 and 71271061, Chongqing Natural Science Foundation (for Distinguished Young Scholars), Grant No. CSCT, 2010BA2003, National High Technology Research and Development Program of China (863 Program), Grant No. 2013AA013801.

\bibliographystyle{elsarticle-harv}
\bibliography{myreference}







\end{document}